%% file: iclr2027_conference.tex
\documentclass{article} 
\usepackage{iclr2027_conference,times}
\usepackage{algorithm}
\usepackage{algpseudocode}
\input{math_commands.tex}

\usepackage{graphicx}
\usepackage{hyperref}
\usepackage{url}
\usepackage{amsmath}
\usepackage{amssymb}
\usepackage{booktabs}
\usepackage{cleveref}
\usepackage{placeins}

\title{OVOGO: Offline Value-Guided On-Policy Generative Optimization for Flow-Based Vision-Language-Action Models}

\author{%
\ignorespaces
\bfseries
Rongzheng Yang$^{1,2}$,
Jiaming Guo$^{1,2}$,
Yuxuan Guo$^{1,3}$,
Zihao Zhang$^{4}$,
Yunkai Gao$^{4}$,
Siming Lan$^{4}$,\\
\bfseries
Shaohui Peng$^{5}$,
Rui Zhang$^{1,2}$,
Tianyun Ma$^{4}$,
Xing Hu$^{1,2}$,
Yunji Chen$^{1,2,4}$\\
$^{1}$State Key Lab of Processors, Institute of Computing Technology,\\
\hspace*{0.7em}Chinese Academy of Sciences, Beijing, China\\
$^{2}$University of Chinese Academy of Sciences\\
$^{3}$University of Science and Technology of China\\
$^{4}$Institute of AI for Industries, Chinese Academy of Sciences\\
$^{5}$Intelligent Software Research Center, Institute of Software,\\
\hspace*{0.7em}Chinese Academy of Sciences, Beijing, China
}

\iclrfinalcopy 
\begin{document}

\maketitle

\begin{abstract}
Flow-based vision-language-action models can benefit from robot experience collected during deployment, but repeatedly acquiring new physical interactions for policy improvement is costly. Offline reinforcement learning provides a natural way to reuse collected experience, while flow-based policies offer an additional opportunity because fresh generation trajectories can be cheaply resampled from the current policy at replay observations. This motivates separating offline value learning from on-policy optimization of the generative process. We propose OVOGO, an offline reinforcement learning framework that realizes this separation for flow-based VLAs. A central challenge in this formulation is to learn a critic that remains grounded in logged robot experience while providing useful value signals for newly generated actions. OVOGO introduces a Return-Anchored Ranking Critic that combines in-sample distributional value learning with guarded local ranking, improving discrimination among generated actions without assigning them artificial return targets. The frozen critic then provides group-relative value signals for selected-transition optimization of fresh action candidates sampled from the current flow policy. Experiments on RoboTwin 2.0, DexJoCo, and LIBERO, as well as a real-robot dexterous manipulation task, show that OVOGO improves the initial VLA across all four settings and achieves higher success rates than the evaluated offline adaptation and value-guided baselines. These results suggest that offline robot experience can be repeatedly reused for value learning while inexpensive generative sampling supports policy optimization, providing an effective approach to experience-based post-training of flow-based VLAs.
\end{abstract}

\section{Introduction}
\label{sec:introduction}

Vision-language-action models (VLAs) have demonstrated strong manipulation capabilities and generalization by combining large-scale, cross-task robot data with pretrained vision-language representations~\cite{kim2024openvla,black2024pi0,physicalintelligence2025pi05}. 
However, further improving these models still relies heavily on collecting high-quality demonstrations, whose acquisition is expensive and difficult to scale. 
In contrast, deployment and autonomous execution naturally produce large amounts of experience spanning successful, failed, and suboptimal behaviors, together with task-outcome signals that behavior cloning cannot fully exploit. 
Reinforcement learning (RL) provides a natural mechanism for learning from such experience, and offline RL is particularly attractive for robotics because it can repeatedly reuse previously collected interactions and their reward feedback without requiring new physical interaction for every policy update~\cite{kostrikov2022iql,physicalintelligence2025recap,wang2026lwd}.
As VLA training moves beyond imitation pretraining toward experience-based post-training, offline RL is emerging as a promising approach for improving large generative VLAs from accumulated robot experience.

Applying offline RL to flow-based VLAs, however, is not straightforward. Because actions are produced through a multi-step flow process, a central question is how value information should be translated into updates of the generative policy. Existing approaches adopt different value-to-policy interfaces. RECAP converts estimated advantages into conditioning variables and extracts an improved policy through supervised training on collected experience~\cite{physicalintelligence2025recap}. This design is simple and scalable, but its improvement supervision is attached to replayed actions rather than actions freshly sampled from the current policy. Other approaches, such as LWD and Q-VGM, use the critic's action gradient to construct improvement targets for the flow dynamics~\cite{wang2026lwd,wang2026qvgm}. These methods enable more direct value-based policy improvement, but make the update depend on reliable action-gradient information from the critic around policy-generated actions. Flow policies offer a complementary opportunity because physical interactions are expensive and should be reused extensively, whereas generation trajectories are inexpensive computational samples that can be repeatedly resampled from the current policy at the same replay observation. This asymmetry suggests that physical experience and generation trajectories need not be consumed at the same rate during policy improvement.

Motivated by this observation, we decouple flow-based VLA reinforcement learning by separating offline value learning from on-policy generative policy optimization. The former learns long-horizon values from a fixed robot dataset, while the latter repeatedly samples fresh actions from the current flow policy at replay observations.
The critic serves as the key interface between these two processes. Directly incorporating policy-generated actions into Bellman updates would make temporal targets depend on value estimates for actions without observed environment outcomes. Errors in these estimates, especially overestimation, can then propagate through bootstrapping and be further exploited during policy improvement.
Keeping Bellman value propagation in-sample provides a more stable foundation for learning long-horizon returns because temporal targets remain grounded in replayed experience. However, restricting all critic supervision to replayed actions would leave the critic unable to reliably distinguish fresh candidates sampled from the evolving policy.
The key challenge is therefore to preserve in-sample temporal value propagation while extending the critic's discriminative ability beyond replayed actions, without compromising the return-grounded value estimates on which offline learning relies.

In this paper, we propose \textbf{OVOGO} (\textbf{O}ffline \textbf{V}alue-Guided \textbf{O}n-\textbf{P}olicy \textbf{G}enerative \textbf{O}ptimization), an offline RL framework for flow-based VLAs that combines stable offline value learning with fresh on-policy optimization of the generative process. For offline value learning, OVOGO introduces a Return-Anchored Ranking Critic (RARC) that separates long-horizon return propagation from local action discrimination. RARC uses in-sample distributional value learning to propagate long-horizon values through real replay transitions without bootstrapping through policy-generated actions, and augments this return-grounded objective with local ranking supervision around replayed actions. Synthetic actions receive only relative preference supervision rather
than artificial Bellman targets, while a guarded ranking objective
combines temperature-dependent gradient decay with a gap cutoff to
control ranking pressure, helping preserve the return scale established
by real experience. With RARC frozen, OVOGO repeatedly samples groups of fresh action candidates from the current flow policy at replay observations and constructs group-relative improvement signals directly from their scalar critic values, without differentiating through the critic. We further adopt a selected-transition flow update inspired by Flash-GRPO~\cite{he2026flashgrpo}, making on-policy generative optimization tractable for large flow-based VLAs.

In our experiments, we evaluate OVOGO on RoboTwin~2.0~\cite{chen2025robotwin}, DexJoCo~\cite{wang2026dexjoco}, and LIBERO~\cite{liu2023libero}, as well as a real-robot dexterous manipulation task. The results demonstrate that OVOGO more effectively improves flow-based VLAs from collected robot experience than existing offline adaptation and value-guided methods, while additional comparisons and ablations analyze the contributions of its critic design and generative policy optimization.

\section{Related Work}

\subsection{Reinforcement Learning for VLA Post-Training}

Recent studies have explored reinforcement learning (RL) as a post-training paradigm for improving pretrained vision-language-action models (VLAs). 
One line of work directly applies on-policy RL to VLAs by repeatedly collecting trajectories from the current policy and optimizing them with PPO~\cite{schulman2017ppo}, GRPO~\cite{shao2024deepseekmath}, or related objectives~\cite{tan2025ript,li2025simplevla,chen2025pirl}.
While effective, such approaches require repeated environment rollouts, which are expensive for robotic systems; consequently, many studies rely primarily on simulated environments for RL training. 
Another line of work seeks to better reuse previously collected robot experience. 
RECAP~\cite{physicalintelligence2025recap} converts learned value information into advantage conditioning and improves the policy through supervised learning, preserving the scalability of imitation-style training but deriving its policy-improvement signal primarily from actions already contained in the collected experience. 
More recent approaches such as LWD~\cite{wang2026lwd} and Q-VGM~\cite{wang2026qvgm} learn explicit critics and use action gradients of the critic to guide flow-policy updates, enabling more direct value-based improvement but placing stronger requirements on the local first-order accuracy of the critic around policy-generated actions. 
In contrast, our method resamples candidate actions from the current flow-based VLA and uses only scalar critic values to construct relative policy-improvement signals, directly optimizing the current generative policy without differentiating through the critic.

\subsection{Policy Optimization for Diffusion and Flow Policies}

RL for diffusion and flow policies has been studied primarily in generic continuous-control settings. 
One class of methods treats the denoising or flow-generation process itself as a sequential policy and directly optimizes the generative trajectory with reinforcement learning~\cite{ren2025dppo,hu2025decisionflow}; another class avoids direct optimization of the full iterative generation process and instead transfers value information through one-step surrogate actors, value-weighted behavior learning, or regularized policy extraction~\cite{park2025fql,tiofack2025guidedflow,liang2026dipole}. 
Most of these approaches are developed for conventional diffusion/flow control policies rather than large-scale VLAs with pretrained vision-language backbones. 
OGPO~\cite{patil2026ogpo} combines an off-policy critic for reusing expensive environment transitions with freshly sampled trajectories from the current generative policy, and our separation between offline value learning and on-policy generative optimization is inspired by this principle. 
OGPO, however, focuses on online fine-tuning of relatively lightweight generative control policies, where the replay buffer is continually refreshed through environment interaction, whereas we study offline post-training of large pretrained VLAs and explicitly address how an offline-trained critic can reliably evaluate actions generated by the evolving policy.
Finally, recent work has shown that single-step policy optimization can substantially reduce the cost of reinforcement learning for large diffusion models~\cite{he2026flashgrpo}; we adopt this idea to make on-policy generative optimization practical for large flow-based VLAs.

\section{Preliminaries}
\label{sec:preliminaries}

\subsection{Flow Matching for VLA Policies}
\label{sec:flow_prelim}

We consider a flow-based VLA policy that generates action chunks conditioned on the current VLA context~\cite{black2024pi0}. Let $s_t$ denote the context at robot timestep $t$, including visual observations, the language instruction, and proprioception. 
We denote by $A_t=[a_t,a_{t+1},\ldots,a_{t+H-1}]$ the action chunk associated with one control transition, where $H$ is the execution horizon.
Following flow matching~\cite{lipman2023flowmatching}, the policy learns a continuous transport from Gaussian noise to the conditional action distribution. Given a data pair $(s_t,A_t)$, Gaussian noise $\epsilon\sim\mathcal{N}(0,I)$, and flow time $\tau\sim\mathcal{U}(0,1)$, we define $x_{t,\tau}=(1-\tau)\epsilon+\tau A_t$, whose target velocity is $A_t-\epsilon$. The action expert $v_\theta(x_{t,\tau},\tau,s_t)$ is trained with
\begin{equation}
\mathcal{L}_{\mathrm{FM}}(\theta)
=
\mathbb{E}_{(s_t,A_t),\,\epsilon,\,\tau}
\left[
\left\|
v_\theta(x_{t,\tau},\tau,s_t)-(A_t-\epsilon)
\right\|_2^2
\right].
\end{equation}

At inference time, sampling starts from $x_t^{[0]}\sim\mathcal{N}(0,I)$ and integrates the learned velocity field over a discretized schedule $0=\tau_0<\tau_1<\cdots<\tau_K=1$. We denote the resulting generation trajectory by $x_t^{[0]},x_t^{[1]},\ldots,x_t^{[K]}$, where $k$ indexes the flow-generation step and $x_t^{[K]}$ represents the generated action chunk $A_t$.

\subsection{Offline Policy Improvement for VLAs}
\label{sec:offline_rl_prelim}

We formulate policy improvement at the action-chunk level. Executing $A_t$ from context $s_t$ produces rewards $r_t,\ldots,r_{t+H-1}$ and leads to the next context $s_{t+H}$. The corresponding discounted chunk return is $
R_t=\sum_{\ell=0}^{H-1}\gamma^\ell r_{t+\ell},$ where $\gamma\in[0,1)$ is the per-step discount factor. The policy objective is to maximize the expected discounted return $J(\pi)=
\mathbb{E}_{\pi}
\left[
\sum_{t\geq0}\gamma^t r_t
\right].$ In the offline setting, policy improvement is performed from a fixed replay dataset
$
\mathcal{D}
=
\left\{
(s_t,A_t,R_t,s_{t+H},d_t)
\right\},$
where $d_t$ indicates termination. The dataset may contain trajectories collected by one or more behavior policies. Offline reinforcement learning aims to improve the policy using only the logged experience in $\mathcal{D}$, without additional environment interaction during optimization. A practical extension is iterative offline policy improvement, where the policy is periodically deployed to collect new experience and the resulting transitions are added to the replay dataset for the next offline optimization round~\cite{physicalintelligence2025recap}. Each optimization round remains offline with respect to its current replay dataset, while the accumulated dataset may expand between rounds.
\section{Method}
\label{sec:method}

OVOGO consists of two stages. First, we train a Return-Anchored Ranking Critic
(RARC) from logged robot experience using return grounded value learning and
local ranking supervision, where synthetic actions provide only relative
ordering constraints and never receive Bellman targets. RARC is then frozen and
used to evaluate fresh action candidates sampled from the current flow policy at
replay observations. Their relative critic values are converted into optimization
signals for selected transition updates of the action expert, without requiring
gradients through the critic.

\begin{figure}[t]
    \centering
    \includegraphics[width=0.93\textwidth]{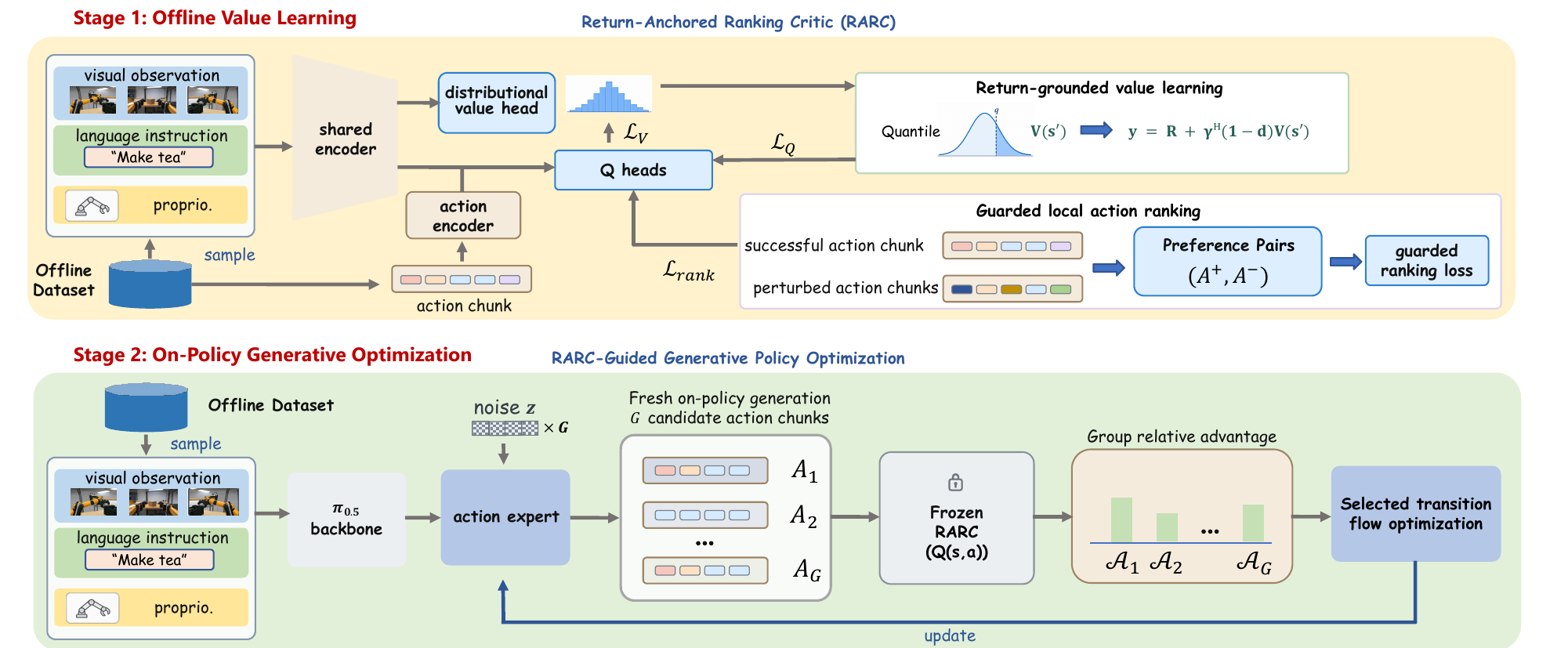}
    \caption{
    \textbf{Overview of OVOGO.} RARC is first trained from logged robot experience
using return-grounded value learning and guarded local ranking. The frozen RARC
then scores fresh policy-generated candidates to guide selected-transition flow
optimization.
    }
    \label{fig:ovogo-overview}
    \vspace{-10pt}
\end{figure}

\subsection{Return-Anchored Ranking Critic}
\label{sec:RARC}

At a replay observation, the actor may generate action chunks that were never
executed by the robot. Their environment outcomes are therefore unavailable,
yet their critic scores determine the direction of the actor update. RARC
separates two sources of supervision. Executed replay transitions establish the
long-horizon return scale, while ranking perturbations provide local ordering
constraints without receiving artificial return targets.

\paragraph{Return grounded value learning.}
Inspired by Distributional Implicit Value Learning (DIVL)~\citep{wang2026lwd}, we
model the distribution of values associated with actions in replay rather
than using a single scalar state value. This representation captures the
heterogeneous value structure in offline robot experience and supports
quantile based value extraction for temporal bootstrapping. RARC contains $M$ ensemble members. Each member $m$ has two scalar action
value heads $Q_{m,1}$ and $Q_{m,2}$ and a categorical state value head
$Z_m$. All members share a multimodal state encoder, while the Q heads
additionally condition on the action chunk. We use $Q_h$,
$h=1,\ldots,2M$, to denote individual Q heads when the member identity is
irrelevant. A superscript $-$ denotes quantities evaluated using the
exponential-moving-average target critic when explicitly indicated.

The categorical value head models the state conditioned distribution of Q
values for actions in the replay dataset. Let
$p_m(c\mid s)=\operatorname{softmax}(Z_m(s))_c$, $c=1,\ldots,C$, denote
the distribution of member $m$ over a fixed categorical support
$\mathcal Z=\{z_1,\ldots,z_C\}$. Within each member, we average the two
target Q estimates as
$\bar Q_m^-(s,A)=
\frac{1}{2}\left(Q_{m,1}^-(s,A)+Q_{m,2}^-(s,A)\right)$.
We use twin Q averaging instead of selecting the smaller estimate as in
clipped double Q learning. This avoids systematically shifting the member
level estimate downward before distributional value fitting. Conservatism in
temporal bootstrapping is handled separately through the value quantile
described below. All scalar Q-head estimates are retained separately and later used for conservative evaluation of policy-generated candidates in
Section~\ref{sec:actor}.

For each $(s_t,A_t)$ sampled from the replay dataset, we project
$\bar Q_m^-(s_t,A_t)$ onto $\mathcal Z$ and use the resulting categorical
distribution to supervise $p_m(\cdot\mid s_t)$
\begin{equation}
\mathcal L_V
=
\mathbb E_{\mathcal D}
\left[
\frac{1}{M}
\sum_{m=1}^{M}
\operatorname{CE}
\left(
\Pi_{\mathcal Z}
\left[
\bar Q_m^-(s_t,A_t)
\right],
p_m(\cdot\mid s_t)
\right)
\right].
\label{eq:rarc_value}
\end{equation}

Following the entropy adaptive value extraction in DIVL, we adjust the
bootstrap quantile according to the entropy of the learned value
distribution. We compute the normalized entropy
$\mathcal H_m(s)=
-\frac{1}{\log C}\sum_{c=1}^{C}
p_m(c\mid s)\log p_m(c\mid s)$
and set
$\tau_m(s)=
\operatorname{clip}
\left(
\tau_{\mathrm{base}}-\alpha\mathcal H_m(s),
\tau_{\min},\tau_{\max}
\right)$.
The corresponding state value is
$V_m(s)=
\operatorname{Quantile}_{\tau_m(s)}
[p_m(\cdot\mid s)]$.
A higher entropy leads to a lower quantile and therefore a more conservative
bootstrap value, while a more concentrated distribution retains a higher
quantile.

Using the chunk level transition defined in Section~\ref{sec:preliminaries}, the
member specific TD target is
\begin{equation}
y_{m,t}
=
R_t+\gamma^H(1-d_t)V_m(s_{t+H}).
\label{eq:rarc_td}
\end{equation}
Both Q heads in member $m$ regress to the same target
\begin{equation}
\mathcal L_Q
=
\mathbb E_{\mathcal D}
\left[
\frac{1}{2M}
\sum_{m=1}^{M}
\sum_{j=1}^{2}
\left(
Q_{m,j}(s_t,A_t)-y_{m,t}
\right)^2
\right].
\label{eq:rarc_q_loss}
\end{equation}

This construction keeps temporal value propagation grounded in replay
transitions with observed rewards and successor states without explicitly
maximizing Q over unexecuted actions. It provides the return grounded value
foundation on which we introduce local action ranking supervision next.

\begin{figure*}[t]
    \centering
    \includegraphics[width=\textwidth]{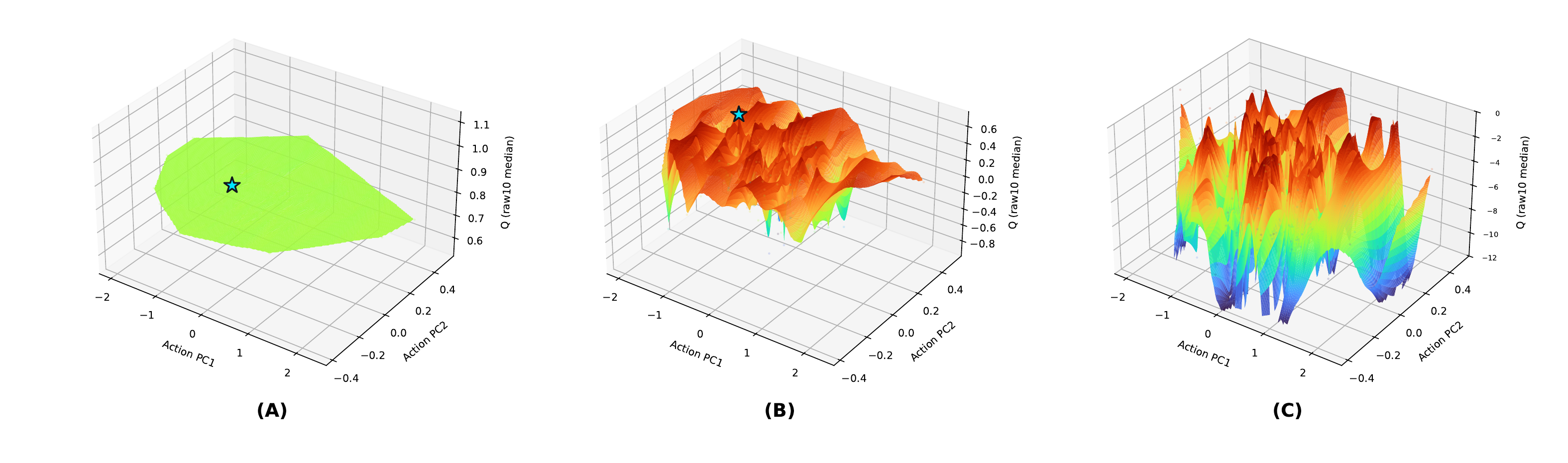}
    \vspace{-30pt}
\caption{
\textbf{Effect of local ranking on the learned action value landscape.}
We visualize the median prediction across ten Q heads over locally sampled
actions obtained through the same forward flow-generation process used for
policy optimization, projected onto their first two principal components.
The star marks the replay action.
\textbf{(A)} Without ranking, the critic shows little local discrimination.
\textbf{(B)} Guarded ranking ($\tau_R=0.1$, $g_{\max}=1$) introduces
clear action discrimination while keeping Q value differences controlled.
\textbf{(C)} Unguarded ranking produces much larger Q value variation.
Raw Q values are shown with intentionally different vertical scales.
}
    \label{fig:rankq_landscape}
    \vspace{-5pt}
\end{figure*}

\paragraph{Guarded local action ranking.}
Return grounded value learning alone may provide insufficient discrimination
among alternative actions at the same replay observation, while excessive
ranking pressure can enlarge synthetic Q gaps beyond the scale established by
replay returns, as illustrated in Figure~\ref{fig:rankq_landscape}.

Inspired by RankQ~\citep{choi2026rankq}, we construct outcome-dependent ranking priors
around replay actions. For an action chunk $A_t$, we sample a Gaussian
direction in normalized action space and construct a mild perturbation
$A_t^\delta$ and a stronger perturbation $A_t^{2\delta}$ along the same
direction. We also construct a uniformly sampled legal action
$A_t^{\mathrm{rand}}$ and a same task mismatched action
$A_t^{\mathrm{perm}}$.

For successful episodes, we use
\[
A_t \succ A_t^\delta,
\qquad
A_t \succ A_t^{2\delta},
\qquad
A_t \succ A_t^{\mathrm{rand}},
\qquad
A_t \succ A_t^{\mathrm{perm}},
\]
\[
A_t^\delta \succ A_t^{2\delta},
\qquad
A_t^{2\delta} \succ A_t^{\mathrm{rand}}.
\]
Successful trajectories provide stronger evidence that their replay actions
are compatible with the task. We therefore use a local smoothness prior under
which smaller perturbations are more likely to preserve useful behavior. The perturbation-based ranking prior is empirically examined in
Section~\ref{sec:critic_analysis}.
The random and mismatched actions provide coarser negative references.
For failed episodes, we only impose
\[
A_t \succ A_t^{\mathrm{rand}}.
\]
Failed trajectories do not provide sufficient evidence for a local quality
ordering among perturbations. This remaining relation encodes only a weak
behavior distribution prior. The replay action comes from the behavior policy
that generated the dataset, while the random action is sampled from the legal
action space. It does not imply that the failed replay action has a high
return.

These preferences provide relative action supervision without assigning
return targets to synthetic actions. A standard pairwise softplus objective
retains nonzero ranking pressure for any finite positive Q gap. Since
synthetic actions have no observed environment outcomes, continued ranking
pressure can push their pairwise differences away from the value scale
anchored by replay returns. For Q head $h$ and a preferred pair
$A^+ \succ A^-$, we define
$g_h=Q_h(s,A^+)-Q_h(s,A^-)$ and use
\begin{equation}
\ell_{\mathrm{rank}}(g_h)
=
\left[
\tau_R
\operatorname{softplus}
\left(
-\frac{g_h}{\tau_R}
\right)
-
\tau_R
\operatorname{softplus}
\left(
-\frac{g_{\max}}{\tau_R}
\right)
\right]_+ .
\label{eq:guarded_rank_loss}
\end{equation}
Here $[x]_+=\max(x,0)$. The temperature $\tau_R$ controls how quickly the
ranking pressure decreases as a positive gap develops. We use a small
temperature to rapidly reduce ranking pressure once a positive gap is
established. The gap cutoff further removes the remaining ranking pressure
once $g_h\geq g_{\max}$.
Together, temperature and the cutoff constitute our guarded ranking
design, controlling the strength and effective range of ranking supervision.
Our tasks use zero intermediate rewards and a terminal reward of one upon
success, which places discounted returns in $[0,1]$. We set the cutoff
threshold $g_{\max}=1$ to match the observed return scale.
The cutoff acts only on the pairwise ranking pressure and does not clip
individual Q values.

We denote the resulting ranking objective by
$\mathcal L_{\mathrm{rank}}$. Synthetic actions participate only in this
objective and never receive TD targets. The complete RARC objective is
\begin{equation}
\mathcal L_{\mathrm{RARC}}
=
\mathcal L_Q
+
\lambda_V\mathcal L_V
+
\lambda_R\mathcal L_{\mathrm{rank}}.
\label{eq:rarc_total_loss}
\end{equation}
Return grounded value learning establishes long-horizon values from replay
transitions, while guarded local action ranking provides the relative action discrimination needed for effective generative policy optimization. The complementary roles of these two objectives are examined through
ablations in Section~\ref{sec:critic_analysis}. Once trained,
RARC remains frozen during the policy optimization stage.

\subsection{RARC-Guided Generative Policy Optimization}
\label{sec:actor}

During actor optimization, observations are sampled from the fixed replay
dataset. At the beginning of each actor iteration, we snapshot the current
policy as the rollout policy $\pi_{\mathrm{old}}$. For each replay observation $s_b$ in the actor batch, the rollout policy generates a group of candidate action chunks from independent Gaussian noise,
\begin{equation}
    \widehat A_{b,i}
\sim
\pi_{\mathrm{old}}(\cdot \mid s_b),
\qquad i=1,\ldots,G.
\end{equation}
The frozen RARC evaluates these candidates and converts their relative values
into policy optimization signals.

\paragraph{Group relative advantage.}
Policy optimization depends on the relative quality of candidates generated
at the same observation. For each Q head, we first center its predictions
within the candidate group,
\begin{equation}
\widetilde q_{h,b,i}
=
Q_h(s_b,\widehat A_{b,i})
-
\frac{1}{G}
\sum_{j=1}^{G}
Q_h(s_b,\widehat A_{b,j}).
\label{eq:centered_q_score}
\end{equation}
This removes the common value offset at the current observation while
preserving the relative value differences among candidates. The categorical
state value head is used only for critic learning and does not serve as the
actor baseline. We then conservatively aggregate the centered scores across Q heads. A
candidate receives a nonzero advantage only when all heads agree on the sign
of its centered score. When all scores are positive, we use the smallest
positive score. When all scores are negative, we use the negative score
closest to zero. A candidate receives zero advantage when the heads disagree
on the improvement direction~\citep{patil2026ogpo}. The resulting
$\mathcal A_{b,i}$ is detached and used as a scalar signal for policy
optimization. This conservative aggregation updates the policy only along relative
improvement directions consistently supported by the critic ensemble, using
the most conservative magnitude among the agreeing estimates. RARC therefore
interfaces with the generative policy only through scalar value feedback, and
policy optimization does not require gradients through the critic.

\paragraph{Selected transition flow optimization.}
RARC evaluates a candidate after its complete action chunk has been generated.
Updating the policy through an entire multistep flow trajectory, however,
introduces substantial computational overhead. We therefore follow the
selected transition optimization of Flash-GRPO~\citep{he2026flashgrpo}.

For each observation $s_b$, we first sample a common generation transition
for the entire candidate group,
$k_b \sim \operatorname{Uniform}\{1,\ldots,K\}$, where transition $k_b$
maps $x^{[k_b-1]}$ to $x^{[k_b]}$. The selected transition uses a
stochastic SDE update, while the remaining generation transitions follow
deterministic ODE dynamics. Each candidate is generated to completion and
receives its advantage $\mathcal A_{b,i}$ from RARC. Policy gradients are
computed only through the transition at $k_b$. All candidates within the
same group share this selected transition, so their relative comparison is
not confounded by differences across flow timesteps.

Let $\rho_{b,i}$ denote the dimension-normalized likelihood-ratio
surrogate between the current policy and the rollout policy at the
selected transition. Different selected timesteps
induce different gradient scales through the SDE discretization. Following
Flash-GRPO, we apply temporal gradient rectification with
$w(k_b)=1/\lambda(k_b)$. The clipped update is
\begin{equation}
\mathcal L_{\mathrm{policy}}
=
-
\mathbb E_{b,i}
\left[
w(k_b)
\min\left(
\rho_{b,i}\mathcal A_{b,i},
\operatorname{clip}\left(
\rho_{b,i},
1-\varepsilon,
1+\varepsilon
\right)
\mathcal A_{b,i}
\right)
\right].
\label{eq:selected_transition_loss}
\end{equation}
Here $\lambda(k_b)$ is the timestep dependent scaling factor induced by the
selected stochastic transition. The exact SDE likelihood, likelihood ratio correction, and form of $\lambda(k_b)$ follow Flash-GRPO and are provided in the Appendix. RARC evaluation of the completed action chunk provides the optimization signal, while the policy gradient is computed through a single selected transition with temporal gradient rectification, retaining the efficiency of single-transition training.

\paragraph{Successful behavior retention.}
In addition to RARC-guided policy optimization, we retain the native
flow-matching objective on successful replay trajectories, denoted by
$\mathcal L^{\mathrm{succ}}$. This auxiliary objective preserves behaviors
in the replay dataset that already achieve task success.

Rather than combining the two objectives with a fixed loss coefficient,
we compute their gradients separately,
$g_{\mathrm{policy}}=\nabla_{\theta}\mathcal L_{\mathrm{policy}}$, $
g_{\mathrm{succ}}
=
\nabla_{\theta}\mathcal L^{\mathrm{succ}},$
and form the actor update as
\begin{equation}
g_{\mathrm{actor}}
=
g_{\mathrm{policy}}
+
\lambda_{\mathrm{eff}} g_{\mathrm{succ}}.
\label{eq:actor_gradient}
\end{equation}
Here $\lambda_{\mathrm{eff}}$ is determined adaptively from the global
norms and cosine similarity of the two gradients, with a cap on the
relative contribution of the successful-behavior gradient. RARC and the remaining VLA backbone stay frozen during policy
optimization, and only the action expert is updated. The exact
successful flow-matching objective, adaptive gradient-capping rule,
and rollout-policy refresh procedure are provided in the Appendix.

\section{Experiments}
\label{sec:experiments}

In this section, we evaluate the effectiveness of OVOGO on flow-based VLA
post-training. In particular, we aim to answer the following research questions:
\textbf{(1)} Compared with existing offline adaptation and value-guided methods, how
effectively can OVOGO improve a pretrained VLA from the same robot experience?
\textbf{(2)} Can RARC provide useful value signals for freshly generated actions, and
does generative policy optimization provide additional gains over inference-time
selection or guidance?
\textbf{(3)} What are the contributions of return grounded value learning,
local ranking, and the guarded ranking design to policy improvement?

\subsection{Experimental Setup}

\paragraph{Benchmarks and data.}
We evaluate on RoboTwin~2.0~\citep{chen2025robotwin},
DexJoCo~\citep{wang2026dexjoco}, and LIBERO~\citep{liu2023libero},
as well as a real-robot dexterous manipulation task.
For each simulation task, the offline dataset contains 100 policy rollout
episodes and 50 demonstration episodes.
Within each simulation benchmark, the actor and critic are
trained jointly across all tasks using one shared model for each.
For RoboTwin, DexJoCo, and LIBERO, we use the publicly released clean50 SFT,
official multi-task $\pi_{0.5}$, and RLinf LIBERO SFT checkpoints,
respectively, both for policy initialization and initial rollout collection.
All methods within each benchmark use the same task split, observation
format, language instructions, action normalization, execution horizon,
and evaluation protocol.

\paragraph{Baselines.}
We compare OVOGO with five baselines.
\textbf{Success BC} adapts the action expert by behavior cloning successful
trajectories from the replay dataset.
\textbf{RECAP} is the value based policy adaptation method introduced with
$\pi_{0.6}^{*}$~\citep{physicalintelligence2025recap}.
We use the Evo-RL implementation~\citep{evorl2026} and fully fine-tune
both the VLM backbone and the action expert.
\textbf{Best-of-N} samples $N$ action chunks from the policy and executes the
candidate with the highest RARC score. We set $N=8$, matching the candidate
budget used by OVOGO, and use the same critic aggregation without updating the
policy.
\textbf{QGF} uses critic gradients to guide the flow generation process at
inference time~\citep{zhou2026qgf}. It starts from the same VLA and uses the offline
trained critic without optimizing the actor from group relative candidate
scores.
\textbf{LWD}~\citep{wang2026lwd} learns a distributional implicit value
function and uses critic action gradients to optimize the flow policy.
We reproduce LWD following its reported algorithm and hyperparameter settings.

\paragraph{Training protocol.}
Our primary experiments follow the standard offline setting. All methods start
from the same initial VLA within each benchmark and use the same fixed
offline dataset without additional environment interaction during optimization.
For OVOGO, the pretrained VLM backbone is frozen and only the action
expert is optimized.
We report the mean and standard deviation of success rates over three
seeds for the repeated simulation experiments.
We additionally evaluate RECAP and OVOGO in an iterative offline setting on
RoboTwin. At round $k$, policy optimization is performed on a fixed accumulated
dataset $\mathcal D^{(k)}$. The updated policy then collects 100 rollout
episodes per task, which are added to the dataset for the following round.
The demonstration episodes remain available throughout all three rounds.
Further implementation and training details are provided in the Appendix.

\subsection{Main Results}

\begin{table}[tp]
\centering
\caption{Task macro success rates (\%) in the standard offline setting.
Results other than the initial policy are reported as mean and standard
deviation over three seeds.}
\label{tab:main_results}
\begin{tabular}{lccc}
\toprule
Method & RoboTwin & DexJoCo & LIBERO \\
\midrule
$\pi_{0.5}$ base
& 60.70 & 27.50 & 71.40 \\
Success BC
& $54.10 \pm 0.36$ & $27.57 \pm 1.90$ & $75.11 \pm 0.45$ \\
RECAP
& $64.23 \pm 1.06$ & $29.03 \pm 3.44$ & $76.11 \pm 0.50$ \\
Best-of-$N$
& $64.23 \pm 1.98$ & $29.73 \pm 0.23$ & $73.97 \pm 1.92$ \\
QGF
& $63.30 \pm 1.48$ & $29.47 \pm 0.45$ & $72.30 \pm 1.11$ \\
LWD
& $67.13 \pm 0.85$ & $32.40 \pm 0.60$ & $75.21 \pm 1.42$ \\
\textbf{OVOGO}
& $\mathbf{68.17 \pm 1.19}$
& $\mathbf{36.90 \pm 0.98}$
& $\mathbf{78.36 \pm 2.00}$ \\
\bottomrule
\end{tabular}
\vspace{-5pt}
\end{table}

\paragraph{Standard offline setting.}
Table~\ref{tab:main_results} reports the performance of all methods under the
standard offline setting described above. OVOGO achieves the highest mean
success rate across all three benchmarks and improves over the initial
policy by 7.47, 9.40, and 6.96 percentage points on RoboTwin, DexJoCo,
and LIBERO, respectively.
RECAP improves over the initial policy on all three benchmarks,
while OVOGO achieves further improvements.
Compared with LWD, which directly optimizes the flow policy
using critic action gradients, OVOGO achieves improvements of 1.04, 4.50,
and 3.15 percentage points, respectively.
The inference-time baselines further examine whether the learned
critic provides useful signals for policy-generated actions.
Both Best-of-$N$ and QGF improve over the initial policy across
all three benchmarks without changing the underlying policy.
OVOGO further improves over both baselines by using critic
feedback to update the generative policy, without requiring
critic evaluation at inference time.

\begin{figure}[tbp]
    \centering
    \includegraphics[width=0.9\linewidth]{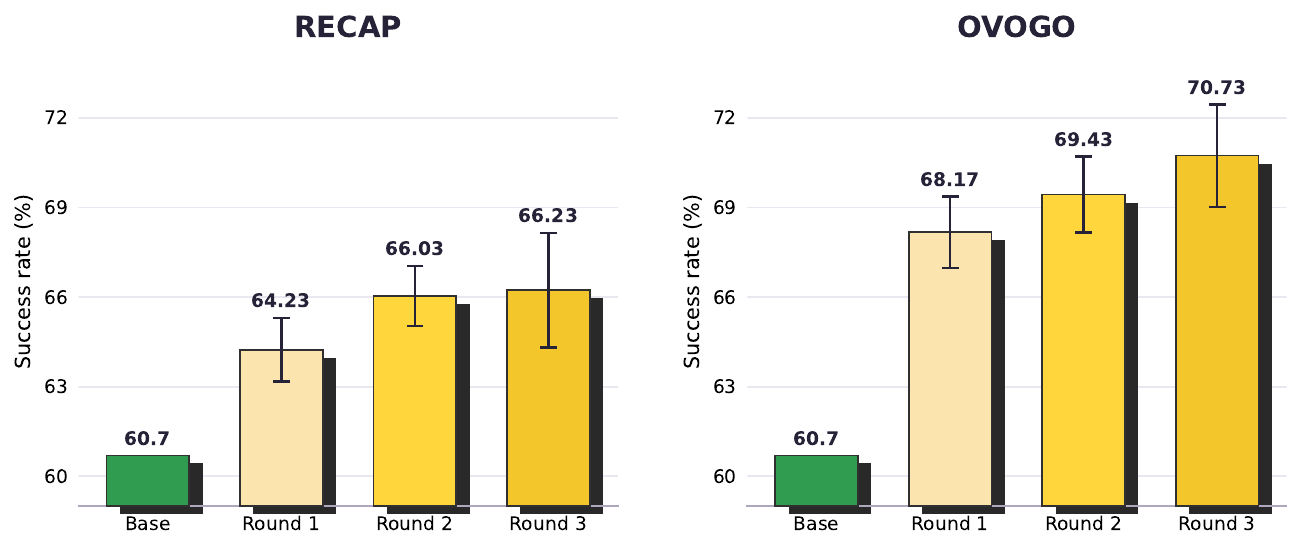}
    \vspace{-5pt}
    \caption{RoboTwin task-macro success rates for RECAP and OVOGO across
    three rounds of offline training and rollout collection.}
    \label{fig:roundwise_results}
    \vspace{-10pt}
\end{figure}

\paragraph{Iterative offline setting.}
We further evaluate RECAP and OVOGO under the iterative offline setting on
RoboTwin. Figure~\ref{fig:roundwise_results} reports the success rates as
newly collected experience is incorporated into the offline dataset
across three rounds. Both methods improve as new experience is incorporated,
while OVOGO achieves a higher mean success rate in each round.

\subsection{Real-Robot Evaluation}
\label{sec:real_robot}

We further evaluate OVOGO on a real-robot block-stacking task using a
UR5e robotic arm equipped with an INSPIRE dexterous hand.
The initial $\pi_{0.5}$ policy is fine-tuned on 50 demonstrations and
is subsequently used to collect 100 rollout episodes for offline
post-training. Each method is evaluated over 20 real-robot trials.

\begin{table}[tbp]
\centering
\caption{Success rates on the real-robot block-stacking task.}
\label{tab:real_robot}
\begin{tabular}{lcc}
\toprule
Method & Successful trials & Success rate (\%) \\
\midrule
$\pi_{0.5}$ base & 9/20 & 45.0 \\
SFT & 11/20 & 55.0 \\
RECAP & 13/20 & 65.0 \\
Best-of-$N$ & 11/20 & 55.0 \\
QGF & 10/20 & 50.0 \\
LWD & 13/20 & 65.0 \\
\textbf{OVOGO} & \textbf{16/20} & \textbf{80.0} \\
\bottomrule
\end{tabular}
\vspace{-5pt}
\end{table}

As shown in Table~\ref{tab:real_robot}, OVOGO improves the success rate
from 45\% to 80\%, outperforming the evaluated baselines on this task.
Compared with RECAP and LWD, which both achieve 65\% success,
OVOGO achieves an additional improvement of 15 percentage points.
These results demonstrate the potential of OVOGO for improving
flow-based VLAs from real robot experience without requiring additional
physical interactions during policy optimization.

\subsection{Analysis of RARC}
\label{sec:critic_analysis}

\begin{table}[tbp]
\centering
\caption{Ablation of return grounded value learning and guarded local
ranking under the standard offline setting.
Results other than the initial policy are reported as mean and standard
deviation over three seeds.}
\label{tab:rankq_ablation}
\small
\begin{tabular}{lcc}
\toprule
Critic objective & RoboTwin (\%) & LIBERO (\%) \\
\midrule
$\pi_{0.5}$ base
& 60.70 & 71.40 \\
Return-grounded only
& $60.97 \pm 1.38$ & $72.44 \pm 2.91$ \\
Guarded ranking only
& $55.90 \pm 1.54$ & $53.26 \pm 4.98$ \\
Return-grounded + ranking ($\tau_R=1$, no cutoff)
& $63.17 \pm 0.76$ & $71.05 \pm 3.10$ \\
Return-grounded + ranking ($\tau_R=0.1$, no cutoff)
& $66.50 \pm 0.56$ & $76.99 \pm 1.04$ \\
\textbf{RARC ($\tau_R=0.1$, $g_{\max}=1$)}
& $\mathbf{68.17 \pm 1.19}$
& $\mathbf{78.36 \pm 2.00}$ \\
\bottomrule
\end{tabular}
\vspace{-5pt}
\end{table}

\paragraph{Contribution of return grounded value learning.}
We examine the complementary roles of return grounded value learning
and local ranking in RARC. All variants use the same fixed offline
dataset and actor optimization procedure.
As shown in Table~\ref{tab:rankq_ablation}, the return-grounded critic
achieves $60.97 \pm 1.38$\% success on RoboTwin, while using ranking
supervision alone without return grounded value learning reduces the
success rate to $55.90 \pm 1.54$\%.
Combining the two objectives in RARC improves the success rate to
$68.17 \pm 1.19$\%.
These results suggest that ranking supervision alone is insufficient
for effective policy improvement, while return grounded value learning
provides a foundation on which local ranking can enhance action
discrimination. A similar pattern is observed on LIBERO, where guarded ranking
alone achieves $53.26 \pm 4.98$\% and the full RARC reaches
$78.36 \pm 2.00$\%.

\paragraph{Effect of temperature and gap cutoff.}
Ranking with $\tau_R=1$ and no cutoff achieves
$63.17 \pm 0.76$\% success on RoboTwin.
Lowering the temperature to $\tau_R=0.1$ without the cutoff increases
the success rate to $66.50 \pm 0.56$\%.
Introducing the gap cutoff at the same temperature further improves
the success rate to $68.17 \pm 1.19$\%.
These results show that both components of guarded ranking contribute
to policy improvement, with the larger gain coming from the reduced
temperature.
The full guarded ranking design improves over ranking with $\tau_R=1$
and no cutoff by 5.00 and 7.31 percentage points on RoboTwin and
LIBERO, respectively. On LIBERO, adding the cutoff at $\tau_R=0.1$ increases the mean
success rate from $76.99 \pm 1.04$\% to $78.36 \pm 2.00$\%.

\paragraph{Validation of the local ranking prior.}
We examine the local ranking prior by applying perturbations at different
scales to the action chunks generated by the initial policy on RoboTwin.
Using the same evaluation initial states and seeds, mild perturbations
with $\delta=0.04$ and stronger perturbations with $2\delta=0.08$
reduce the success rate from 60.7\% to $56.13 \pm 0.80$\% and
$49.40 \pm 6.24$\%, respectively.
The decrease in success as the perturbation magnitude increases
supports the local ranking prior used in RARC.

\paragraph{Critic reliability after policy optimization.}
Since RARC remains frozen during policy optimization, its ability to
evaluate actions generated by the evolving policy is important.
On RoboTwin, Best-of-$N$ using RARC improves the initial policy from
60.7\% to $64.23 \pm 1.98$\%.
Applying the same frozen critic to candidates generated by the final
OVOGO actors further improves their success rate from
$68.17 \pm 1.19$\% to $70.43 \pm 0.65$\%.
These results provide evidence that RARC remains useful for action
selection after policy optimization.
\section{conclusion}

In this work, we present OVOGO, an offline reinforcement learning framework
for improving flow-based vision-language-action models from collected robot
experience. OVOGO learns a return-grounded ranking critic and uses its value
signals to optimize fresh actions generated by the current flow policy.
Experiments on RoboTwin 2.0, DexJoCo, and LIBERO, as well as a real-robot
dexterous manipulation task, show that OVOGO consistently improves
the initial VLA and achieves stronger performance than existing offline
adaptation and value-guided methods. More broadly, OVOGO shows how expensive
physical experience can be repeatedly reused together with inexpensive
generative sampling, providing a scalable path toward experience-driven
post-training of large robot policies.

\subsection*{AI use statement}
In this work, we used generative AI tools to assist with manuscript drafting and code inspection. We did not use these tools to collect experimental observations, generate evaluation outcomes, or produce the numerical results reported in the paper. All figures included in the submitted manuscript were drawn and finalized by the authors; no AI-generated figure was included. Additionally, we used generative AI tools for language polishing, translation, and LaTeX formatting. We reviewed all AI-assisted work, checked technical descriptions and equations against our implementation and cited sources, and checked reported results against experimental records. We take responsibility for the final content of this work, including text and claims produced with the aid of generative AI.

\subsection*{Reproducibility Statement}

To facilitate reproducibility, we provide the implementation code together with the detailed hyperparameter configurations used for training and evaluation. The simulation benchmarks and initial VLA checkpoints used in simulation are publicly available from their original sources. The main algorithmic details, data collection procedures, training procedures, and evaluation settings are described in the paper and supplementary materials. These resources facilitate reproduction of the experimental results presented in this work.

\bibliographystyle{iclr2027_conference}
{
\small
\bibliography{iclr2027_conference}
}

\appendix
\section{Implementation Details}
\label{app:implementation}

This section provides additional implementation details of OVOGO.
We first summarize the complete training procedure in
Algorithm~\ref{alg:ovogo}, and then describe the selected-transition
generative policy optimization, successful-behavior retention, and the
architecture and training configuration of the Return-Anchored Ranking
Critic (RARC). The standard offline setting corresponds to a single
offline round, whereas the iterative offline setting repeats the same
critic--actor optimization procedure after augmenting the replay dataset
between rounds.

\begin{algorithm}[H]
\caption{OVOGO training procedure}
\label{alg:ovogo}
\small
\begin{algorithmic}[1]
\Require Initial replay dataset $\mathcal D^{(1)}$; pretrained VLA
$\pi_{\theta}$; number of offline rounds $R$
\Ensure Adapted policy $\pi_{\theta}$

\For{$r=1,\ldots,R$}
    \Statex \textbf{Stage 1: Train the return-anchored ranking critic}
    \For{each critic update on the fixed dataset $\mathcal D^{(r)}$}
        \State Sample logged transitions from $\mathcal D^{(r)}$
        \State Construct distributional value and Bellman Q targets from logged transitions using the target critic
        \State Construct local ranking pairs around logged actions
        \State Update RARC using
        $\mathcal L_{\mathrm{RARC}}
        =\mathcal L_Q+\lambda_V\mathcal L_V+\lambda_R\mathcal L_{\mathrm{rank}}$
        \State Update the target critic by exponential moving averaging
    \EndFor
    \State Freeze RARC

    \Statex \textbf{Stage 2: Optimize the action expert}
    \State Set $\pi_{\mathrm{old}}\leftarrow\pi_{\theta}$
    \For{each actor update on the fixed dataset $\mathcal D^{(r)}$}
        \State Sample a batch of replay observations
        \For{each replay observation $s_b$ in the batch}
            \State Sample a common $k_b\sim\operatorname{Uniform}\{1,\ldots,K\}$ for the candidate group
            \State Generate $G$ complete chunks under $\pi_{\mathrm{old}}$, with only transition $k_b$ stochastic
            \State Score the completed chunks with the frozen RARC scalar Q heads
            \State Construct detached group-relative advantages using conservative ensemble aggregation
        \EndFor
        \State Compute the selected-transition Flash-GRPO gradient $g_{\mathrm{policy}}$
        \State Compute the success-only flow-matching gradient $g_{\mathrm{succ}}$
        \State Compute $\lambda_{\mathrm{eff}}$ and combine
        $g_{\mathrm{actor}}=g_{\mathrm{policy}}+\lambda_{\mathrm{eff}}g_{\mathrm{succ}}$
        \State Clip $g_{\mathrm{actor}}$ and update only the action expert
        \State Refresh $\pi_{\mathrm{old}}\leftarrow\pi_{\theta}$
    \EndFor

    \If{$r<R$}
        \State Collect rollout episodes $\mathcal E^{(r+1)}$ with the adapted policy
        \State Set $\mathcal D^{(r+1)}\leftarrow
        \mathcal D^{(r)}\cup\mathcal E^{(r+1)}$
    \EndIf
\EndFor
\State \Return $\pi_{\theta}$
\end{algorithmic}
\end{algorithm}

\subsection{Selected-Transition Generative Policy Optimization}
\label{app:selected_transition}

We adopt the selected-transition optimization strategy of
Flash-GRPO~\citep{he2026flashgrpo} to efficiently optimize the
multi-step flow-generation process. The central idea is to evaluate a
completed generated action chunk while propagating policy gradients
through only one stochastic transition of its generation trajectory.

For each replay observation $s_b$, the rollout policy
$\pi_{\mathrm{old}}$ generates a group of $G$ candidate action chunks
from independent Gaussian noise,
\begin{equation}
x_{b,i}^{[0]}\sim\mathcal N(0,I),
\qquad
\widehat A_{b,i}=x_{b,i}^{[K]},
\qquad
i=1,\ldots,G,
\label{eq:app_candidate_sampling}
\end{equation}
where $K$ denotes the number of flow transitions.

Following the iso-temporal grouping principle of
Flash-GRPO, all candidates associated with the same replay observation
share a common selected generation step,
\begin{equation}
k_b
\sim
\operatorname{Uniform}
\{1,\ldots,K\}.
\label{eq:app_selected_step}
\end{equation}
The selected transition maps
$x_{b,i}^{[k_b-1]}$ to $x_{b,i}^{[k_b]}$.
Sharing the same selected timestep within a candidate group avoids
confounding relative candidate comparisons with differences in
generation time.

\paragraph{Stochastic selected transition.}
To follow the notation of Flash-GRPO, we use $t$ for the reverse-time
parameterization in this subsection. This variable is distinct from the
forward flow-time variable $\tau$ used in the flow-matching
preliminaries. Following the SDE formulation used in Flash-GRPO, a
deterministic flow ODE can be transformed into a stochastic transition
while preserving the same marginal generation process. At flow time
$t$, the selected transition is written as
\begin{equation}
\begin{aligned}
\bm{x}_{t-\Delta t}
=
\bm{x}_{t}
+
\Bigg[
\bm{v}_{\theta}(\bm{x}_{t},t)
+
\frac{\sigma_t^2}{2t}
\Big(
\bm{x}_{t}
+
(1-t)\bm{v}_{\theta}(\bm{x}_{t},t)
\Big)
\Bigg]\Delta t
+
\sigma_t\sqrt{\Delta t}\,
\bm{\epsilon},
\end{aligned}
\label{eq:app_flash_sde}
\end{equation}
where
$\bm{\epsilon}\sim\mathcal N(0,I)$,
$\bm{v}_{\theta}$ is the velocity field predicted by the action expert,
and $\sigma_t$ controls the stochasticity of the transition.

The conditional mean of the selected transition is therefore
\begin{equation}
\begin{aligned}
\bm{\mu}_{\theta}(\bm{x}_{t},t)
=
\bm{x}_{t}
+
\Bigg[
\bm{v}_{\theta}(\bm{x}_{t},t)
+
\frac{\sigma_t^2}{2t}
\Big(
\bm{x}_{t}
+
(1-t)\bm{v}_{\theta}(\bm{x}_{t},t)
\Big)
\Bigg]\Delta t .
\end{aligned}
\label{eq:app_flash_mean}
\end{equation}
Accordingly, the selected-transition likelihood is
\begin{equation}
p_{\theta}
\left(
\bm{x}_{t-\Delta t}
\mid
\bm{x}_{t},s_b
\right)
=
\mathcal N
\left(
\bm{\mu}_{\theta}(\bm{x}_{t},t),
\sigma_t^2\Delta t\,I
\right).
\label{eq:app_flash_likelihood}
\end{equation}

All remaining generation steps follow the corresponding deterministic
flow dynamics. Importantly, each candidate is still generated to
completion before being evaluated by the frozen RARC. The critic thus
assigns value to the complete generated action chunk, while the policy
gradient is computed only through the selected transition.

\paragraph{Selected-transition likelihood ratio.}
Let $\ell_{\theta,b,i}$ and $\ell_{\mathrm{old},b,i}$ denote the
log likelihoods assigned to the selected transition under the current
and rollout policies, respectively. Let $D_f$ denote the event
dimension of the selected flow transition. We first form the
dimension-normalized log-likelihood ratio
\begin{equation}
\Delta_{b,i}
=
\frac{
\ell_{\theta,b,i}
-
\ell_{\mathrm{old},b,i}
}{
D_f
}.
\label{eq:app_log_ratio}
\end{equation}
For numerical stability, the normalized log ratio is clipped before
exponentiation,
\begin{equation}
\rho_{b,i}
=
\exp
\left[
\operatorname{clip}
\left(
\Delta_{b,i},
-c_{\rho},
c_{\rho}
\right)
\right],
\label{eq:app_policy_ratio}
\end{equation}
where $c_{\rho}$ is the log-ratio clipping threshold listed in
Table~\ref{tab:hyperparameters}. We use dimension normalization and
log-ratio clipping as numerical stabilization for the
high-dimensional selected transition.

\paragraph{Temporal gradient rectification.}
A stochastic SDE transition introduces a timestep-dependent scaling
factor into the policy gradient. Following Flash-GRPO, this factor is
\begin{equation}
\lambda(t)
=
\frac{\sqrt{\Delta t}}{\sigma_t}
+
\frac{
\sigma_t\sqrt{\Delta t}(1-t)
}{
2t
}.
\label{eq:app_lambda_t}
\end{equation}
Without correction, different selected flow timesteps can therefore
produce substantially different gradient magnitudes even when their
advantages have comparable scales. We apply temporal gradient
rectification using
\begin{equation}
w(t)
=
\frac{1}{\lambda(t)}.
\label{eq:app_tgr_weight}
\end{equation}

Let $\mathcal A_{b,i}$ denote the detached group-relative advantage
provided by RARC. The selected-transition policy objective is
\begin{equation}
\begin{aligned}
\mathcal L_{\mathrm{policy}}
=
-
\mathbb E_{b,i}
\Bigg[
w(t_{k_b})
\min
\Big(
&
\rho_{b,i}\mathcal A_{b,i},
\\
&
\operatorname{clip}
(
\rho_{b,i},
1-\varepsilon,
1+\varepsilon
)
\mathcal A_{b,i}
\Big)
\Bigg].
\end{aligned}
\label{eq:app_selected_policy_loss}
\end{equation}

Only the selected transition contributes policy-likelihood gradients.
The remaining generation trajectory is used to construct the complete
candidate action but is not differentiated through. Similarly, RARC is
frozen during actor optimization, and the critic-derived advantage is
detached before the policy update.

The rollout policy $\pi_{\mathrm{old}}$ is refreshed from the current
action expert after each policy update. This maintains a locally
consistent reference distribution for the PPO-style likelihood ratio.

\subsection{Successful-Behavior Retention and Gradient Control}
\label{app:successful_retention}

RARC-guided policy optimization encourages the action expert to shift
probability toward candidates preferred by the critic. At the same
time, we retain behaviors in the offline replay dataset that already
achieve task success by applying the native flow-matching objective to
successful trajectories.

Let $\mathcal D_{\mathrm{succ}}$ denote successful replay samples.
For a successful logged action chunk $A_t$, we sample
$\epsilon\sim\mathcal N(0,I)$ and flow time
$\tau\sim\mathcal U(0,1)$ and construct
\begin{equation}
x_{\tau}
=
(1-\tau)\epsilon+\tau A_t.
\label{eq:app_success_path}
\end{equation}
The corresponding successful-behavior flow-matching objective is
\begin{equation}
\mathcal L_{\mathrm{FM}}^{\mathrm{succ}}
=
\mathbb E_{
(s_t,A_t)\sim\mathcal D_{\mathrm{succ}},
\,\epsilon,\tau
}
\left[
\left\|
v_{\theta}(x_{\tau},\tau,s_t)
-
(A_t-\epsilon)
\right\|_2^2
\right].
\label{eq:app_success_fm}
\end{equation}

Rather than combining the two objectives through a fixed scalar loss
weight, we compute their gradients separately. Let
\begin{equation}
g_{\mathrm{policy}}
=
\nabla_{\theta}
\mathcal L_{\mathrm{policy}},
\qquad
g_{\mathrm{succ}}
=
\nabla_{\theta}
\mathcal L_{\mathrm{FM}}^{\mathrm{succ}}.
\label{eq:app_actor_gradients}
\end{equation}

A fixed auxiliary coefficient can allow the flow-matching gradient to
dominate the critic-guided policy update when the two objectives have
substantially different gradient scales. We therefore use a nominal
coefficient $\lambda_0$ together with an adaptive gradient cap.

Define the global gradient norms
\begin{equation}
a=\|g_{\mathrm{policy}}\|_2,
\qquad
b=\|g_{\mathrm{succ}}\|_2,
\label{eq:app_grad_norms}
\end{equation}
and, when both gradients have nonzero norm, their cosine similarity
\begin{equation}
c
=
\frac{
\left\langle
g_{\mathrm{policy}},
g_{\mathrm{succ}}
\right\rangle
}{
ab
}.
\label{eq:app_grad_cosine}
\end{equation}
For numerical stability, $c$ is clipped to the interval $[-1,1]$.

Let $\rho_{\mathrm{BC}}\in(0,1)$ denote the target maximum contribution
of the weighted successful-behavior gradient relative to the norm of
the combined actor gradient. We compute
\begin{equation}
\xi
=
\frac{
\rho_{\mathrm{BC}}^2c
+
\rho_{\mathrm{BC}}
\sqrt{
\max\left(
0,\,
1-\rho_{\mathrm{BC}}^2
+
\rho_{\mathrm{BC}}^2c^2
\right)
}
}{
1-\rho_{\mathrm{BC}}^2
},
\label{eq:app_bc_xi}
\end{equation}
followed by
\begin{equation}
\lambda_{\mathrm{cap}}
=
\frac{
\xi a
}{
b+\epsilon_g
},
\label{eq:app_bc_cap}
\end{equation}
and
\begin{equation}
\lambda_{\mathrm{eff}}
=
\min
\left(
\lambda_0,
\max(0,\lambda_{\mathrm{cap}})
\right).
\label{eq:app_lambda_eff}
\end{equation}

The final actor gradient is then
\begin{equation}
g_{\mathrm{actor}}
=
g_{\mathrm{policy}}
+
\lambda_{\mathrm{eff}}
g_{\mathrm{succ}}.
\label{eq:app_actor_gradient}
\end{equation}

For non-degenerate gradients, this construction enforces the target cap
\begin{equation}
\frac{
\|
\lambda_{\mathrm{eff}}
g_{\mathrm{succ}}
\|_2
}{
\|
g_{\mathrm{policy}}
+
\lambda_{\mathrm{eff}}
g_{\mathrm{succ}}
\|_2
}
\leq
\rho_{\mathrm{BC}}.
\label{eq:app_bc_fraction}
\end{equation}
Thus, $\rho_{\mathrm{BC}}$ is an upper-bound target rather than a fixed
mixture coefficient. When the successful-behavior gradient is small,
the nominal coefficient $\lambda_0$ is retained. When it becomes large,
its effective weight is reduced so that the auxiliary update does not
dominate the RARC-guided policy gradient.

When $b\leq\epsilon_g$, we retain the nominal coefficient
$\lambda_0$. If $a=0$ while $b>\epsilon_g$, the adaptive rule yields
$\lambda_{\mathrm{eff}}=0$. These cases avoid unstable scaling when one
of the two gradient components is numerically negligible.

Only successful replay trajectories contribute to this auxiliary
flow-matching objective. Failed trajectories do not receive behavior
cloning supervision, and we do not introduce an additional
action-space mean-squared-error cloning objective. The two gradient
components are explicitly combined before gradient clipping and a
single optimizer step.

\subsection{RARC Critic Architecture and Training}
\label{app:rarc_architecture}

RARC uses frozen pretrained visual and language encoders together with
a lightweight trainable critic. The language backbone is
Gemma~3 270M, and the vision backbone is SigLIP2-So400M with
$224\times224$ visual inputs. The pretrained Gemma~3 and SigLIP2
parameters remain frozen during critic optimization; only the
critic-specific projection layers and prediction heads are trained.
The trainable critic core and encoder projection contain approximately
$7.78$M parameters.

For RoboTwin, the critic receives three visual observations
corresponding to the base camera and the two wrist cameras, together
with the language instruction and robot proprioceptive state. The same critic design is used for DexJoCo, LIBERO, and
the real-robot experiments with their respective observation
and action interfaces.

RARC contains $M=5$ ensemble members. Each member contains two scalar
action-value heads and one categorical state-value head, yielding ten
raw scalar Q estimates in total. The categorical value head uses
$C=201$ atoms on the fixed support $[-0.1,1.1]$. We retain the ten Q
estimates separately when evaluating policy-generated candidates.

\paragraph{EMA target critic.}
For training stability, we maintain an exponential-moving-average copy
of the complete critic, including the trainable encoder projection,
scalar Q heads, and categorical value heads. The target Q predictions
are used to construct the distributional value supervision, while the
categorical value distribution used for TD bootstrapping is evaluated
with the EMA target critic. The target critic is updated after every
critic optimization step with coefficient
$\eta_{\mathrm{tgt}}=0.005$.

The return-grounded distributional value objective, entropy-adaptive
quantile extraction, scalar Q objective, and guarded local ranking
objective are defined in Section~\ref{sec:RARC}; we do not repeat those
formulations here. Their implementation-specific numerical settings are
summarized in Table~\ref{tab:hyperparameters}.

\section{Experimental Details}
\label{app:experimental_details}

\subsection{Environments and Policy Initialization}
\label{app:environments}

\paragraph{RoboTwin 2.0.}
We evaluate on ten tasks from RoboTwin~2.0
\citep{chen2025robotwin}.
Each task uses visual observations, a language instruction, and robot
proprioception. The policy predicts action chunks of length $H=50$.

We initialize the policy from the publicly released
$\pi_{0.5}$ clean50 checkpoint.\footnote{
\url{https://huggingface.co/madokalif/pi05-robotwin2-clean50-sft}
}

The RoboTwin environment action representation has dimension
$50\times14$, while the $\pi_{0.5}$ action expert internally operates
on a $50\times32$ flow representation before conversion to environment
actions.

\paragraph{DexJoCo.}
We evaluate on eleven tasks from DexJoCo
\citep{wang2026dexjoco}, including five bimanual tasks and six
single-arm tasks. The multi-task policy predicts action chunks of
length $H=30$. Single-arm and bimanual tasks share a unified
multi-task policy interface using the benchmark-specific state and
action representations.

We initialize from the official multi-task $\pi_{0.5}$
checkpoint.\footnote{
\url{https://huggingface.co/DexJoCo/DexJoCo-Pi05/tree/main/pi05_dexjoco_multi_task}
}
The released checkpoint is provided in JAX/Orbax format. We convert
the checkpoint to PyTorch before training, including the
vision-language and action-expert parameters and the corresponding
normalization statistics. The converted model is used consistently
for all DexJoCo comparisons.

\paragraph{LIBERO.}
We evaluate on 40 tasks from LIBERO~\citep{liu2023libero},
covering four task suites: LIBERO-Spatial, LIBERO-Object,
LIBERO-Goal, and LIBERO-Long (LIBERO-10), with ten tasks in
each suite.
The policy predicts a 10-step action chunk and executes the
first five actions before the next policy call.

We initialize the policy from the publicly released RLinf
$\pi_{0.5}$ LIBERO SFT checkpoint.\footnote{
\url{https://huggingface.co/RLinf/RLinf-Pi05-LIBERO-SFT}
}
The same checkpoint is used for all LIBERO comparisons
and for collecting the initial policy rollout episodes.

Except for benchmark-specific observation and action interfaces, we
use the same OVOGO training configuration across the three simulation
benchmarks.

\subsection{Task Suites}
\label{app:task_suites}

Table~\ref{tab:task_suite} lists the tasks used in RoboTwin and DexJoCo
together with a short description of the required manipulation. The
descriptions summarize the task semantics indicated by the official
benchmark task definitions and language instructions.

\begin{table*}[t]
\centering
\caption{Task suites used for RoboTwin~2.0 and DexJoCo.}
\label{tab:task_suite}
\small
\resizebox{\textwidth}{!}{%
\begin{tabular}{lllp{0.48\textwidth}}
\toprule
Benchmark & Task & Type & Task description \\
\midrule
RoboTwin 2.0
& \texttt{adjust\_bottle}
& Bimanual
& Grasp the bottle with the side-appropriate arm and move it to the corresponding elevated target side. \\
&
\texttt{click\_alarmclock}
& Bimanual
& Press the top button of the alarm clock with the arm corresponding to its location. \\
&
\texttt{click\_bell}
& Bimanual
& Press the top of the bell with the arm corresponding to its location. \\
&
\texttt{dump\_bin\_bigbin}
& Bimanual
& Lift the small tabletop bin and pour its contents into the large bin. \\
&
\texttt{handover\_mic}
& Bimanual
& Grasp the microphone with one arm and hand it over to the opposite arm. \\
&
\texttt{open\_laptop}
& Bimanual
& Grasp the laptop lid and open it beyond the required hinge angle. \\
&
\texttt{pick\_dual\_bottles}
& Bimanual
& Grasp the two bottles and move them to their corresponding left and right elevated targets. \\
&
\texttt{place\_a2b\_right}
& Bimanual
& Grasp object A, place it to the right of object B, and release it. \\
&
\texttt{place\_object\_stand}
& Bimanual
& Grasp the target object, place it on the display stand, and release it. \\
&
\texttt{press\_stapler}
& Bimanual
& Press down on the stapler with the arm corresponding to its location. \\
\midrule
DexJoCo
& \texttt{bimanual\_assembly}
& Bimanual
& Grasp the tray with the left hand and the peg with the right hand, then insert the peg into the hole. \\
&
\texttt{bimanual\_hanoi}
& Bimanual
& Execute the final two Tower-of-Hanoi moves: medium disk middle-to-right, then small disk left-to-right. \\
&
\texttt{bimanual\_microwave\_cook}
& Bimanual
& Open the microwave, place the food inside, close the door, and press the start button. \\
&
\texttt{bimanual\_photograph}
& Bimanual
& Hold the camera with the left hand, align it with the logo, and press the shutter with the right hand. \\
&
\texttt{bimanual\_unlock\_ipad}
& Bimanual
& Grasp the iPad and enter the password 123 to unlock it. \\
&
\texttt{click\_mouse}
& Single-arm
& Move the mouse to the purple mouse pad and click the left mouse button. \\
&
\texttt{fold\_glasses}
& Single-arm
& Fold the glasses and place them into the case. \\
&
\texttt{hammer\_nail}
& Single-arm
& Use the hammer to drive the nail into the wooden board. \\
&
\texttt{pick\_bucket}
& Single-arm
& Place the boxed food into the bucket and then lift the bucket. \\
&
\texttt{pinch\_tongs}
& Single-arm
& Grasp the tongs and perform three consecutive open-close motions. \\
&
\texttt{water\_plant}
& Single-arm
& Grasp the watering can and apply water to the plant. \\
\bottomrule
\end{tabular}
}
\end{table*}
\FloatBarrier

\paragraph{LIBERO task suites.}
We use all ten tasks from each of the four LIBERO task suites.
LIBERO-Spatial focuses on manipulation under varying spatial
relationships. LIBERO-Object varies the objects to be manipulated.
LIBERO-Goal examines different task objectives under fixed
object and spatial configurations. LIBERO-Long contains more
complex long-horizon manipulation tasks involving multiple steps.

\subsection{Training Hyperparameters}
\label{app:hyperparameters}

Table~\ref{tab:hyperparameters} summarizes the principal OVOGO
hyperparameters. Unless otherwise stated, the same optimization
configuration is used across the simulation benchmarks.

\begin{table}[t]
\centering
\caption{Main hyperparameters used in OVOGO.}
\label{tab:hyperparameters}
\small
\begin{tabular}{lll}
\toprule
Component & Hyperparameter & Value \\
\midrule
RARC & Ensemble members $M$ & $5$ \\
     & Q heads per member & $2$ \\
     & Total scalar Q heads & $10$ \\
     & Value support atoms $C$ & $201$ \\
     & Value support range & $[-0.1,1.1]$ \\
     & Discount factor $\gamma$ & $0.995$ \\
     & $\tau_{\mathrm{base}}$ & $0.7$ \\
     & Entropy coefficient $\alpha_{\mathrm{ent}}$ & $0.2$ \\
     & Minimum quantile $\tau_{\min}$ & $0.5$ \\
     & Maximum quantile $\tau_{\max}$ & $0.7$ \\
     & Target EMA $\eta_{\mathrm{tgt}}$ & $0.005$ \\
     & Value-loss weight $\lambda_V$ & $1.0$ \\
     & Ranking-loss weight $\lambda_R$ & $1.0$ \\
     & Optimizer & AdamW \\
     & Learning rate & $1\times10^{-4}$ \\
     & Weight decay & $1\times10^{-4}$ \\
     & Global batch size & $32$ \\
     & Gradient-norm clipping & $5$ \\
\midrule
Ranking & Perturbation scale $\delta$ & $0.04$ \\
        & Ranking temperature $\tau_R$ & $0.1$ \\
        & Gap cutoff threshold $g_{\max}$ & $1.0$ \\
\midrule
Actor & Optimizer & Adafactor \\
      & Learning rate & $1\times10^{-4}$ \\
      & State batch size & $8$ \\
      & Candidates per state $G$ & $8$ \\
      & Candidate microbatch size & $16$ \\
      & Flow transitions $K$ & $10$ \\
      & PPO clipping $\varepsilon$ & $0.01$ \\
      & Log-ratio clipping $c_{\rho}$ & $20$ \\
      & Gradient-norm clipping & $1$ \\
\midrule
Retention & Nominal FM coefficient $\lambda_0$ & $0.005$ \\
          & Gradient fraction $\rho_{\mathrm{BC}}$ & $0.2$ \\
          & Numerical constant $\epsilon_g$ & $10^{-12}$ \\
\bottomrule
\end{tabular}
\end{table}
\FloatBarrier

\subsection{Baseline Configurations}
\label{app:baseline_configs}

All policy-training baselines start from the same pretrained policy
and use the same initial offline dataset as OVOGO.
For the policy-training methods in simulation, policy optimization
runs for 10,000 steps, and we evaluate the best saved checkpoint.

\paragraph{Success BC.}
Success BC performs supervised adaptation using successful
trajectories from the same offline replay dataset. We optimize the
native flow-matching objective of $\pi_{0.5}$ while keeping the
pretrained vision-language backbone frozen and updating the action
expert. Failed replay trajectories do not contribute to the Success
BC objective. We use a learning rate of $1\times10^{-5}$ and a
global batch size of 256.

\paragraph{RECAP.}
We implement RECAP~\citep{physicalintelligence2025recap} using the
Evo-RL implementation~\citep{evorl2026}. RECAP starts from
the same pretrained policy and the same initial replay
dataset as OVOGO. We follow the Evo-RL training configuration
for advantage-conditioned policy adaptation and fully fine-tune
both the vision-language backbone and the action expert.
We use a learning rate of $2\times10^{-5}$ and a batch size of 64.
For comparable value representations, we use the same Gemma~3
270M language encoder and SigLIP2-So400M vision encoder with
$224\times224$ visual inputs as in RARC.

\paragraph{Best-of-$N$.}
Best-of-$N$ is an inference-time selection baseline and does not update
the policy parameters. At each observation, the policy generates
$N=8$ candidate action chunks. The candidates are evaluated using the
same frozen RARC used by OVOGO, and the candidate with the highest
aggregated critic score is executed. Using the same candidate budget
as OVOGO isolates the benefit of policy optimization from the benefit
of critic-based candidate selection.

\paragraph{QGF.}
QGF is an inference-time value-guidance baseline. We use the same
pretrained policy and the same RARC critic as OVOGO. During flow
generation, action gradients of the critic are used to steer the
generation trajectory toward higher-value actions, while the actor
parameters themselves remain unchanged. This comparison separates
inference-time value guidance from parameter-space policy
optimization.

\paragraph{LWD.}
LWD combines distributional implicit value
learning with critic-guided optimization of the flow policy.
Unlike OVOGO, which uses scalar critic values to construct
group-relative policy optimization signals, LWD uses critic action
gradients to construct policy improvement targets.
As the official training code was unavailable, we reproduce LWD
following the algorithmic details and hyperparameter settings
reported in the original paper.
LWD uses the same initial policy and offline replay dataset as
OVOGO within each benchmark.

\subsection{Offline Data and Training Protocol}
\label{app:data_protocol}

We distinguish the standard offline setting used for the main
comparison from the iterative offline experiment conducted on
RoboTwin.

\paragraph{Standard offline setting.}
For the main comparison, all methods are trained from the same fixed
offline dataset and receive no additional environment interactions
during optimization.

RoboTwin contains ten tasks. For each task, the initial replay dataset
contains 100 policy rollout episodes and 50 demonstration episodes.
The complete initial RoboTwin dataset therefore contains 1000 policy
rollout episodes and 500 demonstration episodes.

DexJoCo contains eleven tasks. Each task likewise contains 100 policy
rollout episodes and 50 demonstration episodes, resulting in 1100
policy rollout episodes and 550 demonstration episodes in total.

LIBERO contains 40 tasks across four task suites.
Each task contains 100 policy rollout episodes and 50 demonstration
episodes, resulting in 4000 policy rollout episodes and 2000
demonstration episodes in total.

The initial rollout episodes are collected using the corresponding
pretrained policies described in Section~\ref{app:environments}.
Their initial task-macro success rates are 60.7\%, 27.5\%, and
71.40\% on RoboTwin, DexJoCo, and LIBERO, respectively.

Within the standard offline setting, the replay dataset remains fixed
throughout critic learning and actor optimization. Fresh candidate
actions generated by the flow policy are used only to construct
policy-improvement signals. They are not treated as new environment
transitions, are not assigned synthetic rewards, and are not inserted
into the replay dataset.

\paragraph{Iterative offline setting on RoboTwin.}
We additionally evaluate OVOGO and RECAP under an iterative offline
protocol on RoboTwin. Let $\mathcal D^{(k)}$ denote the replay dataset
available in round $k$. Critic and actor optimization within that
round use only the fixed dataset $\mathcal D^{(k)}$.

After completing the offline optimization stage, the updated policy is
deployed to collect 100 new rollout episodes for each of the ten
RoboTwin tasks, giving 1000 new rollout episodes per round. We denote
this newly collected experience by $\mathcal E^{(k+1)}$ and construct
the replay dataset for the following round as
\begin{equation}
\mathcal D^{(k+1)}
=
\mathcal D^{(k)}
\cup
\mathcal E^{(k+1)}.
\label{eq:app_iterative_replay}
\end{equation}
The original demonstration trajectories remain available throughout
all rounds.

This protocol preserves the offline nature of each optimization round:
new environment experience is collected only between rounds, whereas
policy-generated candidates sampled during optimization never enter
the replay dataset.

\subsection{Evaluation Protocol}
\label{app:evaluation_protocol}

For every simulation benchmark, we evaluate each task independently
using 100 evaluation episodes. For task $j$, the episode success rate is
\begin{equation}
\operatorname{SR}_j
=
\frac{
N^{\mathrm{succ}}_j
}{
N^{\mathrm{eval}}_j
},
\end{equation}
where $N^{\mathrm{succ}}_j$ and $N^{\mathrm{eval}}_j$ denote the number
of successful and total evaluation episodes for task $j$,
respectively.

The benchmark-level metric reported in the main paper is the
task-macro success rate. For a benchmark containing $J$ tasks,
\begin{equation}
\operatorname{MacroSR}
=
\frac{1}{J}
\sum_{j=1}^{J}
\operatorname{SR}_j.
\label{eq:app_macro_success}
\end{equation}
Thus, every task contributes equally to the final score regardless of
differences in its individual success rate.

We compute the task-macro success rate for each seed and report its
mean and standard deviation over three seeds for the repeated
simulation experiments. The initial policy is evaluated directly
from the corresponding pretrained checkpoint without policy adaptation.

All compared methods within the same benchmark use the same task
split, language instructions, observation processing, action
normalization, policy initialization, and evaluation protocol.
RoboTwin and DexJoCo policies generate and execute 50-step and
30-step action chunks, respectively, while LIBERO policies
generate 10-step action chunks and execute the first five actions.

For Best-of-$N$, the only additional inference-time computation is the
generation and critic evaluation of $N=8$ candidates. For QGF, the
critic is used only for inference-time gradient guidance. For OVOGO,
evaluation uses the optimized policy directly without critic-based
candidate selection.

\subsection{Real-Robot Experimental Details}
\label{app:real_robot}

\paragraph{Hardware and observations.}
Our real-robot platform consists of a Universal Robots UR5e arm
with six revolute joints and a five-finger INSPIRE Robots
RH56DFX-series dexterous hand with six actuator channels.
A fixed external Orbbec Femto Bolt camera observes the workspace,
and a wrist-mounted Intel RealSense L515 provides a close-up view.
The policy receives RGB images from both cameras, resized to
$224\times224$ pixels, together with the language instruction and
the current robot state.

\paragraph{State and action representation.}
The proprioceptive state and the executed action each have 12
dimensions. The state contains the measured joint and actuator
positions, while the action specifies absolute position targets
in the same channel order.
The first six dimensions correspond to the UR5e arm joint positions
in radians. The remaining six dimensions correspond to the little
finger, ring finger, middle finger, index finger, thumb flexion,
and thumb rotation of the dexterous hand, represented in raw
actuator counts.

At each policy call, $\pi_{0.5}$ predicts an action chunk of length
$H=32$. The internal flow representation has dimension $32\times32$,
and the first 12 channels at each step form the $32\times12$
action chunk passed to the robot controller.
The policy executes 32 actions before the next policy call under
the default execution setting.

\paragraph{Initial policy and rollout collection.}
We collect 50 demonstrations of the block-stacking task and use them
to fine-tune $\pi_{0.5}$ for 30,000 optimizer steps with a global
batch size of 256.
We use AdamW with $\beta_1=0.9$, $\beta_2=0.95$,
$\epsilon=10^{-8}$, and weight decay of $10^{-10}$.
The gradient norm is clipped to 1.0.
The learning rate warms up linearly for 1,000 steps to
$5\times10^{-5}$ and follows a cosine decay schedule with a
final learning rate of $5\times10^{-6}$.

The checkpoint at step 30,000 serves as the initial policy for
all real-robot comparisons. This policy is also used to collect
100 additional rollout episodes for offline post-training.

\paragraph{Evaluation.}
Each method is evaluated on 20 real-robot trials.
We report the number of successful trials and the corresponding
empirical success rate in Table~\ref{tab:real_robot}.
The demonstration-trained initial policy and the subsequent
SFT baseline are reported separately.

\subsection{Compute Platform}
\label{app:compute}

All experiments are implemented in PyTorch.
We use four NVIDIA A100 GPUs for distributed OVOGO critic and actor
training.

During critic optimization, the Gemma~3 270M language backbone and the
SigLIP2-So400M vision backbone with $224\times224$ inputs remain frozen.
Only the critic-specific projection and prediction modules are updated.
During actor optimization, RARC and the pretrained vision-language
backbone remain frozen, and only the $\pi_{0.5}$ action expert is
optimized.

The benchmark-specific policy checkpoints and the DexJoCo
JAX/Orbax-to-PyTorch conversion are described in
Section~\ref{app:environments}; we do not repeat the checkpoint URLs
here.
\end{document}

%% file: math_commands.tex
\usepackage{amsmath,amsfonts,bm}

\def\eqref#1{equation~\ref{#1}}

\def\1{\bm{1}}

\DeclareMathAlphabet{\mathsfit}{\encodingdefault}{\sfdefault}{m}{sl}
\SetMathAlphabet{\mathsfit}{bold}{\encodingdefault}{\sfdefault}{bx}{n}

